\documentclass[10pt,twocolumn,letterpaper]{article}

\usepackage{cvpr}              
\usepackage{multirow}
\usepackage[ruled,linesnumbered]{algorithm2e}
\usepackage{cuted}

\usepackage[dvipsnames]{xcolor}

\definecolor{cvprblue}{rgb}{0.21,0.49,0.74}
\usepackage[pagebackref,breaklinks,colorlinks,citecolor=cvprblue]{hyperref}

\def\paperID{****} 
\def\confName{CVPR}
\def\confYear{2024}

\title{FlowDA: Unsupervised Domain Adaptive Framework for Optical Flow Estimation}

\author{Miaojie Feng \quad\quad Longliang Liu \quad\quad Hao Jia \quad\quad Gangwei Xu \quad\quad Xin Yang\footnotemark[1]\\
{\normalsize School of EIC, Huazhong University of Science and Technology}\\
{\tt\small \{fmj, longliangl, haojia, gwxu, xinyang2014\}@hust.edu.cn}
}

\begin{document}
\maketitle
\begin{abstract}

Collecting real-world optical flow datasets is a formidable challenge due to the high cost of labeling. A shortage of datasets significantly constrains the real-world performance of optical flow models. Building virtual datasets that resemble real scenarios offers a potential solution for performance enhancement, yet a domain gap separates virtual and real datasets. This paper introduces FlowDA, an unsupervised domain adaptive (UDA) framework for optical flow estimation. FlowDA employs a UDA architecture based on mean-teacher and integrates concepts and techniques in unsupervised optical flow estimation. Furthermore, an Adaptive Curriculum Weighting (ACW) module based on curriculum learning is proposed to enhance the training effectiveness. Experimental outcomes demonstrate that our FlowDA outperforms state-of-the-art unsupervised optical flow estimation method SMURF by 21.6\%, real optical flow dataset generation method MPI-Flow by 27.8\%, and optical flow estimation adaptive method FlowSupervisor by 30.9\%, offering novel insights for enhancing the performance of optical flow estimation in real-world scenarios. The code will be open-sourced after the publication of this paper.

\end{abstract}

{
\renewcommand{\thefootnote}{\fnsymbol{footnote}}
\footnotetext[1]{Corresponding author.}
}    
\section{Introduction}
\label{sec:intro}

\begin{figure}[t]
    \centering
    \includegraphics[width=0.95\linewidth]{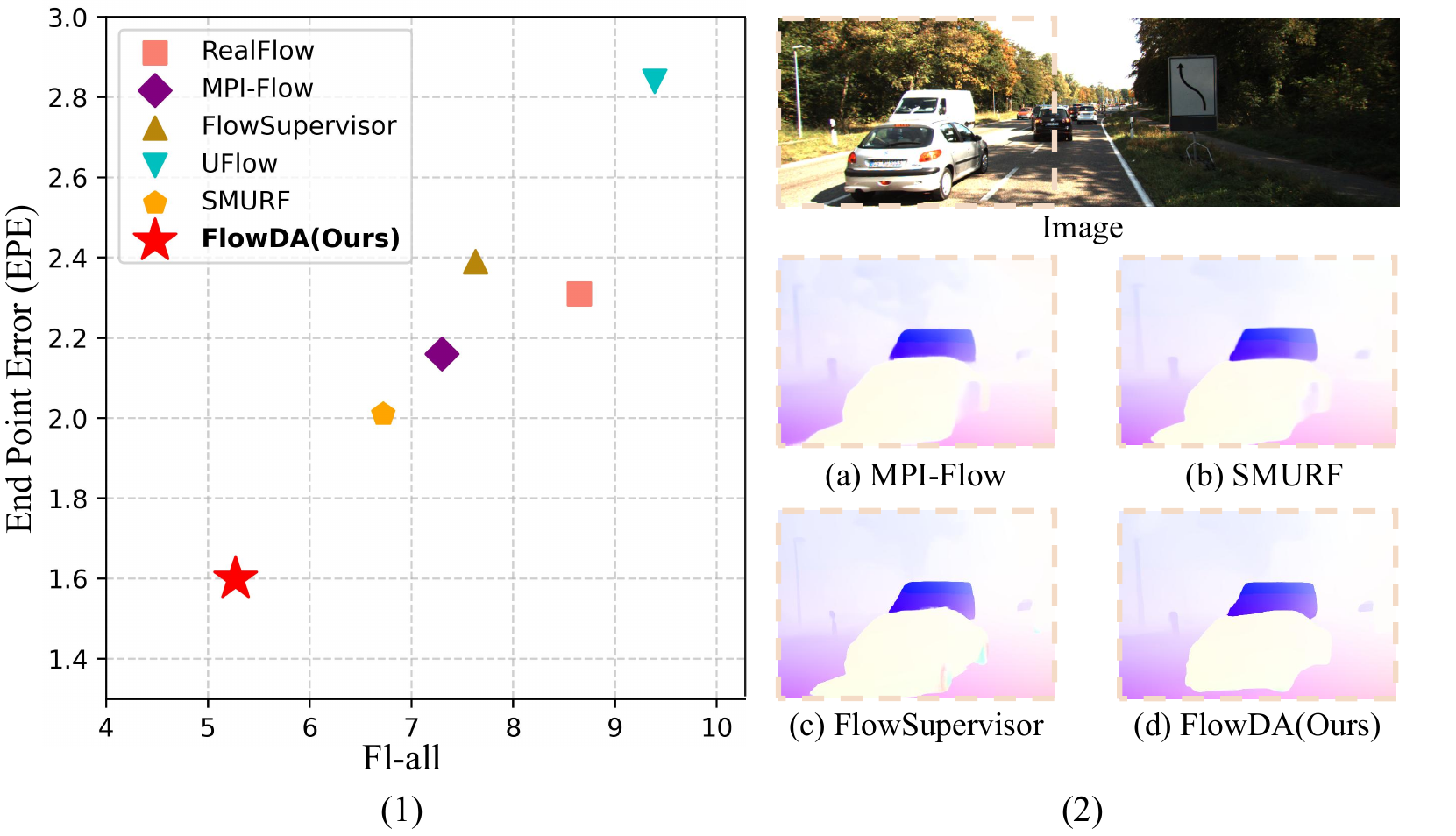}
    \vspace{-10pt}
    \caption{(1) Quantitative comparison (End Point Error vs. Fl-all on KITTI-2015). Our proposed method FlowDA, shown in the red star, achieves state-of-the-art performance compared with other methods. (2) Qualitative comparison.}
    \label{fig:comparison}
    \vspace{-15pt}
\end{figure}

\hspace{1em} Optical flow estimation is an important task in the field of computer vision, which aims to capture the motion information of pixels in an object or scene between successive frames. Applications of optical flow estimation include object tracking, autonomous driving, and motion analysis.

Due to the high cost of acquiring optical flow labels in the real world, real-scene optical flow datasets with ground truth are very scarce. For example, the commonly used publicly available real-world dataset of automated driving with dynamic objects is only KITTI-2015~\cite{menze2015joint}, and there are only 200 pairs of training images. The scarcity of authentic datasets is the performance bottleneck of optical flow estimation algorithms in the real world. To enhance the performance of optical flow estimation methods in practical settings, previous research has primarily focused on three categories: (1) unsupervised methods~\cite{stone2021smurf,liu2019ddflow,jonschkowski2020matters,liu2019selflow,liu2020learning,luo2021upflow,li2023gyroflow+}, (2) generating real optical flow datasets~\cite{aleotti2021learning,han2022realflow,liang2023mpi}, and (3) utilizing virtual datasets~\cite{xiong2023stereoflowgan,im2022semi}. Despite not relying on labels, unsupervised methods encounter noise in their constructed optimization targets, particularly when photometric consistency is not established. This noise represents the primary factor limiting the performance of such methods. The methods of generating the real optical flow datasets attempt to construct the dataset of the real scene by rendering the second frame image from the first frame image and optical flow. These methods frequently depend on depth estimation networks or segmentation networks, with the effectiveness and applicability constrained by the auxiliary networks. Compared with the above methods, leveraging virtual datasets appears as a viable solution to enhance optical flow network performance in diverse real-world scenarios. Virtual datasets offer the advantage of providing entirely accurate optical flow labels. Meanwhile, advances in graphics processing technology have significantly simplified the generation of various virtual datasets. Nonetheless, a domain gap arises between synthetic and real datasets, leading to performance degradation in models trained on synthetic data when applied to actual scenarios. The research on this problem in optical flow estimation is relatively lacking. Previous work~\cite{im2022semi} proposes a semi-supervised method that consists of parameter separation and a student output connection. However, due to the instability of pseudo-labels and the inadequate integration of the characteristics of the optical flow estimation, the final performance is significantly inferior to the first two methods. Overall, the domain adaptive training framework for optical flow estimation has still received insufficient attention, and the full potential of virtual datasets remains untapped.

This paper introduces FlowDA, an unsupervised adaptive framework for optical flow estimation that transfers the optical flow model from the virtual data domain to the real data domain. FlowDA employs a UDA architecture based on mean-teacher and integrates concepts and techniques in unsupervised optical flow estimation. Specifically, FlowDA consists of a teacher model and a student model. The teacher model generates pseudo optical flow labels for unlabeled real data and its parameters are updated using an exponential moving average (EMA). On this basis, a series of improvements are integrated. (1) In order to enhance the reliability of the generated pseudo-labels, FlowDA incorporates the crop self-training technique (employ optical flow estimated on the complete images to self-train optical flow estimated on cropped images) in unsupervised optical flow estimation. This practice is rooted in the fact that optical flow estimated on the complete images provides more reliable supervision for cropped images, especially in moving regions beyond the boundary~\cite{liu2019ddflow}. Different from self-training using the same model~\cite{stone2021smurf} or two models trained separately~\cite{liu2019ddflow} in unsupervised optical flow estimation, FlowDA is based on mean-teacher architecture and updates the teacher model through EMA to improve the stability of pseudo-labels. (2) In order to eliminate inaccuracies in the pseudo-labels, FlowDA employs occlusion detection based on forward-backward consistency~\cite{liu2019selflow}, as the inaccurate area is typically the occluded region. (3) In addition to the conventional pseudo-label loss and supervised loss, FlowDA mines the supervision information of the target domain itself through the photometric consistency loss, which is the core of the unsupervised optical flow estimation methods. (4) Additionally, we introduce an Adaptive Curriculum Weighting module (ACW) based on curriculum learning~\cite{bengio2009curriculum,wang2021survey} to facilitate better model convergence. The ACW module dynamically adjusts the weight of loss on different pixels according to the learning difficulty, allowing the network to learn progressively from simple to complex.

As shown in Figure~\ref{fig:comparison}, FlowDA fully harnesses the potential of virtual datasets, outperforming the most advanced unsupervised optical flow estimation method SMURF~\cite{stone2021smurf} by 21.6\%, real optical flow dataset generation method MPI-Flow~\cite{liang2023mpi} by 27.8\%, and adaptive method for optical flow estimation FlowSupervisor~\cite{im2022semi} by 30.9\%, demonstrating the reliability of building virtual datasets to improve the accuracy of real-world optical flow estimation. At the same time, FlowDA exhibits reliable versatility across various weather conditions and scenarios. Overall, the main contributions of this paper are:

\begin{itemize}[leftmargin=*]
    \vspace{0pt}
    \item We introduce FlowDA, an unsupervised domain adaptive framework for optical flow estimation, which employs a UDA architecture based on mean-teacher and integrates techniques in unsupervised optical flow estimation.
    \vspace{0pt}
    \item We propose the Adaptive Curriculum Weighting (ACW) module, facilitating model convergence by progressively selecting effective and prioritized supervisory signals.
    \vspace{0pt}
    \item FlowDA optimally leverages the potential of virtual datasets, surpassing the most advanced unsupervised optical flow estimation methods, dataset generation methods, and other adaptive methods for optical flow estimation, providing an effective example for improving the performance of optical flow estimation in real scenarios.
    \vspace{0pt}
\end{itemize}

\section{Related Work}
\label{sec:related work}

\subsection{Optical Flow Estimation}
\hspace{1em} Optical flow estimation, as a classic problem in the field of computer vision, has been studied for decades. Traditional methods~\cite{horn1981determining,lucas1981iterative,farneback2003two} are mainly based on the assumption of pixel intensity variation and consistency. However, their performance is constrained when dealing with complex scenes, non-rigid motion, occlusions, and rapid dynamic changes. Deep learning has revolutionized optical flow estimation, with three stages of mainstream architecture evolution: encoder-decoder~\cite{dosovitskiy2015flownet,ilg2017flownet,cheng2017segflow,xiang2018deep,zhai2019optical}, pyramid~\cite{sun2018pwc,sun2019models,hui2018liteflownet,hui2020lightweight,hui2020liteflownet3,zhao2020maskflownet}, and iterative optimization~\cite{teed2020raft,jiang2021learning,sun2022skflow,zhang2021separable,xu2021high,zhao2022global,huang2022flowformer,shi2023videoflow}. In addition, some methods~\cite{xu2022gmflow,lu2023transflow} attempt to estimate optical flow by global matching.

\begin{figure*}[t]
    \centering
    \includegraphics[width=0.90\linewidth]{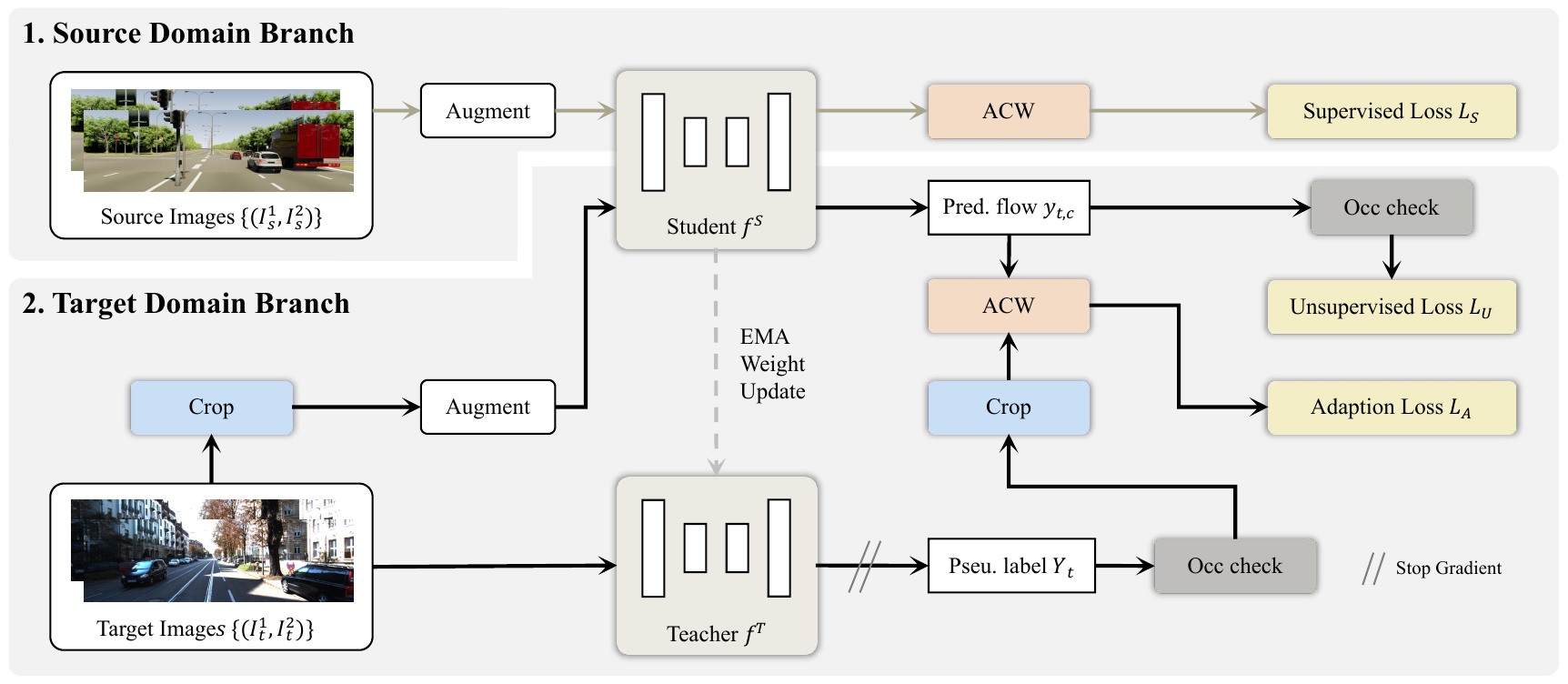}
    \vspace{-5pt}
    \caption{Overall framework of FlowDA. FlowDA includes a teacher model and a student model, with the teacher model generating pseudo optical flow labels for unlabeled real data. The student model is jointly optimized under the constraints of source domain supervised loss $L_{S}$, adaptation loss $L_{A}$, and unsupervised loss $L_{U}$. Meanwhile, the parameters of the teacher model are updated using an exponential moving average (EMA).}
    \label{fig:overall}
    \vspace{-15pt}
\end{figure*}

While the supervised optical flow estimation method has made significant progress, it encounters a pervasive challenge in practical applications: significant performance degradation. This challenge primarily arises from the inherent difficulty of acquiring real-world optical flow labels. To enhance the performance of optical flow estimation methods in practical scenarios, prior research has primarily been divided into three categories: (1) unsupervised methods~\cite{stone2021smurf,liu2019ddflow,jonschkowski2020matters,liu2019selflow,liu2020learning,luo2021upflow,li2023gyroflow+}, (2) generation of real optical flow datasets~\cite{aleotti2021learning,han2022realflow,liang2023mpi}, and (3) domain adaptation through virtual datasets~\cite{xiong2023stereoflowgan,im2022semi}. Unsupervised methods SMURF~\cite{stone2021smurf} integrates the state-of-the-art supervised optical flow estimation architecture into unsupervised methods, significantly enhancing the performance. Nonetheless, the optimization target constructed by unsupervised methods is inherently noisy, and relying solely on this optimization constrains the performance of such methods. Dataset generation methods aim to generate the second frame image from the first frame image and optical flow. MPI-Flow~\cite{liang2023mpi} constructs a layered depth representation from single-view images, employing the camera matrix and plane depths to compute optical flow for each plane, resulting in the generation of new view images. However, such methods often require dependence on depth estimation or segmentation networks, limiting their applicability to specific scenes and influencing the resultant effects. Utilizing virtual datasets appears as a viable solution to enhance the performance of optical flow networks in diverse real-world scenarios. StereoFlowGAN~\cite{xiong2023stereoflowgan} mitigates the domain gap by altering the style of synthetic and real data, and its effectiveness is primarily constrained by the performance of the style transfer network. FlowSupervisor~\cite{im2022semi} proposes a semi-supervised method that consists of parameter separation and a student output connection. However, due to the instability of the pseudo-label and the lack of integration of the characteristics of the optical flow estimation, its performance is not ideal. The domain adaptive training framework has received inadequate attention, and the potential of virtual datasets has not been fully realized. Our research further explores the transition from the virtual to the real domain in optical flow estimation and offers fresh perspectives for enhancing optical flow estimation performance in real-world settings.

\subsection{Unsupervised Domain Adaptation}

\hspace{1em} In Unsupervised Domain Adaptation (UDA), a model trained on a labeled source domain undergoes adaptation to an unlabeled target domain. UDA methods are typically categorized into adversarial training~\cite{gong2021dlow,tsai2018learning,vu2019advent} and self-training approaches~\cite{hoyer2023mic,hoyer2022daformer,sakaridis2018model,yang2020fda,zou2018unsupervised}. In adversarial training, a domain discriminator, learned within the framework of Generative Adversarial Networks (GANs), encourages domain-invariant inputs, features, or outputs. In self-training, the network is trained using pseudo-labels~\cite{lee2013pseudo} from the target domain. Unsupervised domain adaptation has witnessed extensive exploration in the fields of image classification, semantic segmentation, and object detection. In optical flow estimation, research on unsupervised domain adaptation is relatively scarce~\cite{xiong2023stereoflowgan,im2022semi}. Our FlowDA uses mean-teacher-based self-training methods, combined with insights and techniques from unsupervised optical flow estimation, and is substantially ahead of previous works.

\section{Method}
\label{sec:method}

\hspace{1em} The primary objective of this study is to optimize the utilization of virtual datasets and propose a reliable approach to improve the accuracy of optical flow models in real-world scenes. In this section, we will provide a detailed explanation of FlowDA's comprehensive framework, training strategy, and loss function.

\subsection{Overall Framework of FlowDA}

\hspace{1em} Given image pairs $\left \{ (I_{s}^{1}, I_{s}^{2}) \right \}$ from the source domain (virtual data) with optical flow labels $\left \{ Y_{s} \right \}$ and image pairs $\left \{ (I_{t}^{1}, I_{t}^{2}) \right \}$ from the target domain (real data), the objective of FlowDA is to leverage both the source and target domain data in order to enhance the performance of the optical flow estimation model on the target domain, where the source domain has optical flow labels while the target domain does not. In other fields, such as classification, several strategies have been proposed to address domain gap, which can be categorized into adversarial training~\cite{gong2021dlow,tsai2018learning,vu2019advent} and self-training~\cite{hoyer2023mic,hoyer2022daformer,sakaridis2018model,yang2020fda,zou2018unsupervised} methods. In this research, we employ the mean-teacher-based self-training method, as it is known to be more stable compared to adversarial training. However, it is not advisable to directly apply domain adaptive methods used in classification and other fields to optical flow estimation, as they do not adequately explore the distinct constraints and training strategies associated with this task. Valuable insights and ideas from unsupervised optical flow estimation, such as photometric consistency loss, occlusion detection, crop self-training, etc., can provide important contributions. Consequently, we propose FlowDA, which combines the mean-teacher-based unsupervised domain adaptive architecture with concepts and techniques of unsupervised optical flow estimation to realize domain adaptation. Moreover, we propose an Adaptive Curriculum Weighting (ACW) module based on curriculum learning to assist in model training to obtain better convergence results.

The overall framework of FlowDA is shown in Figure~\ref{fig:overall}. FlowDA utilizes two models, the teacher model $f^{T}$ and the student model $f^{S}$. The teacher model $f^{T}$ generates pseudo labels for the student model $f^{S}$, and its weights $\phi$ are the exponential moving average (EMA) of the weights $\theta$ of $f^{S}$ with smoothing factor $\lambda$:
\begin{equation}
    \phi_{n+1}\longleftarrow \lambda \phi_{n} + (1-\lambda )\theta_{n}
    \label{eqn:ema}
\end{equation}
where $n$ denotes a training step. FlowDA is comprised of two parts: the source domain branch and the target domain branch. We will introduce the overall framework of FlowDA from the two branches, and introduce the Adaptive Curriculum Weighting module and the loss function respectively in Sections~\ref{subsec:acw}-\ref{subsec:loss}.

\noindent\textbf{Source domain branch.} The source domain branch uses images $(I_{s}^{1}, I_{s}^{2})$ of the source domain (virtual dataset) and optical flow label $Y_{s}$ for supervised training of the student model. After data augmentation, $I_{s}^{1}$ and $I_{s}^{2}$ are input into the student optical flow estimation network $f^{S}$ to obtain the predicted optical flow $y_{s}$. The ACW module selects a region mask $M_{s}$ with low learning difficulty, based on the predicted optical flow $y_{s}$ and optical flow label $Y_{s}$. In calculating the supervised loss $L_{S}$ of $y_{s}$ and $Y_{s}$, different weights are given according to the mask $M_{s}$.

\noindent\textbf{Target domain branch.} The target domain branch is trained in a self-training and unsupervised manner using the data from the target domain. Images pair $(I_{t}^{1},I_{t}^{2})$ is fed to the teacher model $f^{T}$ and the student model $f^{S}$. Differently, the input images to the teacher model $f^{T}$ undergo no data augmentation, and the resulting optical flow $Y_{t}$ serves as the pseudo-label for the student model. Meanwhile, the images input to the student model are randomly cropped to obtain $(I_{t,c}^{1},I_{t,c}^{2})$, and during the cropping process, the cropping parameters $\psi$ (position and frame size) are recorded. After data augmentation, $I_{t,c}^{1}$ and $I_{t,c}^{2}$ are input into $f^{S}$ to obtain the predicted optical flow $y_{t,c}$. The occlusion detection module (‘Occ check’ in Figure~\ref{fig:overall}), based on forward-backward consistency~\cite{liu2019selflow}, derives occluded regions $O_{t}$ from the output of the teacher model. Subsequently, $Y_{t}$ and $O_{t}$ are cropped based on the cropping parameters $\psi$, resulting in pseudo-label $Y_{t,c}$ and the corresponding occlusion mask $O_{t,c}$. The ACW module outputs mask $M_{t}$ with low learning difficulty based on the predicted optical flow $y_{t,c}$ and optical flow label $Y_{t,c}$. In calculating the adaptation loss $L_{A}$, different weights are given according to the mask $M_{t}$. In addition, predicted optical flow $y_{t,c}$ and cropped images $(I_{t,c}^{1},I_{t,c}^{2})$ are used to construct unsupervised loss $L_{U}$.

\begin{figure*}[t]
    \centering
    \includegraphics[width=0.9\linewidth]{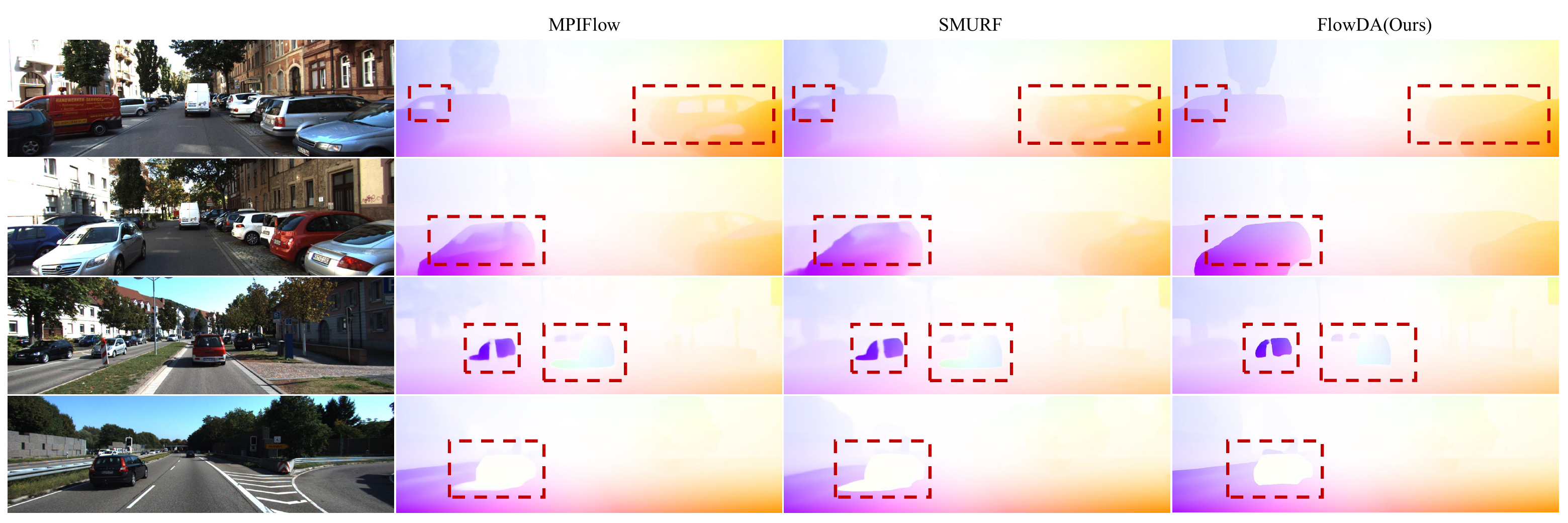}
    \vspace{-5pt}
    \caption{Visualization of flow predictions on KITTI. The second to fourth columns represent the results of MPI-Flow~\cite{liang2023mpi}, SMURF~\cite{stone2021smurf} and our FlowDA, respectively. Red boxes highlight obvious improvements achieved by our method.}
    \label{fig:vis_kitti}
    \vspace{-10pt}
\end{figure*}

\subsection{Adaptive Curriculum Weighting Module}
\label{subsec:acw}

\hspace{1em} Curriculum learning~\cite{bengio2009curriculum} is a machine-learning approach that aims to enhance model performance by incrementally increasing the difficulty or complexity of the training data. Inspired by self-paced curriculum learning~\cite{kumar2011learning}, we introduce an Adaptive Curriculum Weighting Module to facilitate network convergence. This module quantifies the learning difficulty by calculating the End Point Error (EPE) per pixel and is employed in the source and target domain branches.

In the source domain branch, we compute the EPE between the predicted optical flow $y_{s}$ and optical flow label $Y_{s}$, followed by calculating the mean $\mu$ and standard deviation $\sigma$. Regions with an EPE less than or equal to $\mu+N\cdot\sigma$ are classified as simple and easy to learn, while regions with an EPE greater than $\mu+N\cdot\sigma$ are regarded as more challenging, where $N$ is a hyperparameter. When calculating the loss function $L_{S}$, a larger weight is used for areas with low learning difficulty, and a smaller weight is used for difficult areas. Gradually amplifying the value of $N$ during the training process leads to a progressive increase in learning difficulty, thus achieving the desired effect of curriculum learning.

In the target domain branch, due to the usual inaccuracy of the pseudo-label $Y_{t,c}$ in the occluded region $O_{t,c}$, we stop the gradient within it. The ACW module operates on the unoccluded region, and the calculation process is similar to the source domain branch, with inputs $y_{t,c}$ and $Y_{t,c}$.

\subsection{Loss Function}
\label{subsec:loss}

\hspace{1em} FlowDA optimizes the student network $f^{S}$ using three loss functions: supervised loss $L_{S}$ in the source domain branch, adaptation loss $L_{A}$, and unsupervised loss $L_{U}$ in the target domain branch.

\noindent\textbf{Source domain supervised loss $L_{S}$.} Given the output $y_{s}$ of the student model, the optical flow label $Y_{s}$, and the area mask $M_{s}$ with low learning difficulty, the supervised loss $L_{S}$ is calculated as follows:
\begin{equation}
    L_{S}=\varepsilon_{1}\cdot M_{s} \odot L_{1}(y_{s},Y_{s})+\varepsilon_{2}\cdot (1-M_{s})\odot L_{1}(y_{s},Y_{s})
    \label{eqn:sup}
\end{equation}
\noindent{where $L_{1}(\cdot,\cdot)$ denotes the loss of L1 norm, $\odot$ represents element-wise multiplication, $\varepsilon_{1}$ and $\varepsilon_{2}$ are the weights of low and high learning difficulty regions, respectively.}

\noindent\textbf{Adaptation loss $L_{A}$.} Given the output optical flow $y_{t,c}$, pseudo-label $Y_{t,c}$, occlusion mask $O_{t,c}$, and area mask $M_{t}$ with low learning difficulty, the adaptation loss is calculated as follows:
\begin{equation}
    \begin{split}
        L_{A}= & O_{t,c}\odot (\varepsilon_{1}\cdot M_{t}\odot L_{1}(y_{t,c},Y_{t,c}) + \\
        & \varepsilon_{2}\cdot (1-M_{t})\odot L_{1}(y_{t,c},Y_{t,c})) 
    \end{split}
    \label{eqn:ada}
\end{equation}

\noindent\textbf{Unsupervised loss  $L_{U}$.} The unsupervised loss $L_{U}$ consists of the smoothness constraint $L_{smooth}$ and photometric consistency loss $L_{photo}$:
\begin{equation}
    L_{U}=L_{smooth}+O \odot L_{photo}
    \label{eqn:photo}
\end{equation}
\noindent{where $O$ is the occlusion mask obtained based on the forward-backward consistency detection. Specifically, we employ a first-order smoothness constraint~\cite{jonschkowski2020matters,tomasi1998bilateral} and an SSIM-based photometric consistency loss~\cite{wang2004image,ranjan2019competitive}.}

The total loss function is defined as follows:
\begin{equation}
    L_{total}=\alpha \cdot L_{S} + \beta \cdot L_{A} + \gamma \cdot L_{U}
\end{equation}
where $\alpha$, $\beta$, and $\gamma$ are hyperparameters, and we set them to 1 by default in the experiment.

\section{Experiments}
\label{sec:experiments}

\subsection{Dataset}
\textbf{Flyingchairs}~\cite{dosovitskiy2015flownet} and \textbf{Flyingthings}~\cite{mayer2016large} are both synthetic datasets generated by randomly moving foreground objects over a background image. As a standard practice, we use ‘C’ and ‘T’ to represent the two datasets, respectively.

\noindent\textbf{VKITTI 2}~\cite{cabon2020virtual} is a synthetic dataset for autonomous driving scenarios. The dataset is synthesized using Unreal Engine and contains videos generated from different virtual urban environments. We use ‘V’ to refer to this dataset.

\noindent\textbf{KITTI-2012}~\cite{geiger2012we} and \textbf{KITTI-2015}~\cite{menze2015joint} are datasets for real-world autonomous driving scenarios and benchmarks for optical flow estimation. KITTI-2015 is more challenging because it includes dynamic objects. KITTI-2015 has a multi-view extension (4,000 training and 3,989 test) dataset without ground truth. We refer to the training and test sets for multi-view extension as Ktrain and Ktest for short, respectively.

\noindent\textbf{GHOF}~\cite{li2023gyroflow+} is a real dataset for optical flow and homography estimation. The GHOF dataset comprises a collection of videos with gyroscope data, recorded using smartphones. Data acquisition covers various environments and seasons, including regular scenes (RE), low-light scenes (Dark), winter scenes with fog (Fog), summer scenes with rain (Rain), and snowy mountain scenes (Snow). The optical flow is annotated using~\cite{liu2008human}. The GHOF training set includes approximately 10,000 images without optical flow labels and the evaluation set consists of 530 pairs of images.

\subsection{Implementation Details}
\hspace{1em} For the optical flow estimation module, we select RAFT~\cite{teed2020raft} which represents state-of-the-art architecture for supervised optical flow. We train RAFT using official implementation without any modifications and use its pre-trained weights on Flyingchairs~\cite{dosovitskiy2015flownet} and Flyingthings~\cite{mayer2016large} to fine-tune on VKITTI~\cite{cabon2020virtual} (30k iterations, batch size of 6). Unless otherwise specified, we initialize the student model and the teacher model using the weights fine-tuned on VKITTI. The default smoothing factor $\lambda$ for updating the teacher model is set to 0.999. The parameter ‘$N$’ in the ACW module starts at 1 and increases linearly to 5. Weights $\varepsilon _{1}$ and $\varepsilon _{2}$ are 0.9 and 0.1, respectively. For all of our experiments, we employ the AdamW~\cite{loshchilov2017decoupled} optimizer.

\begin{figure}[t]
    \centering
    \includegraphics[width=1\linewidth]{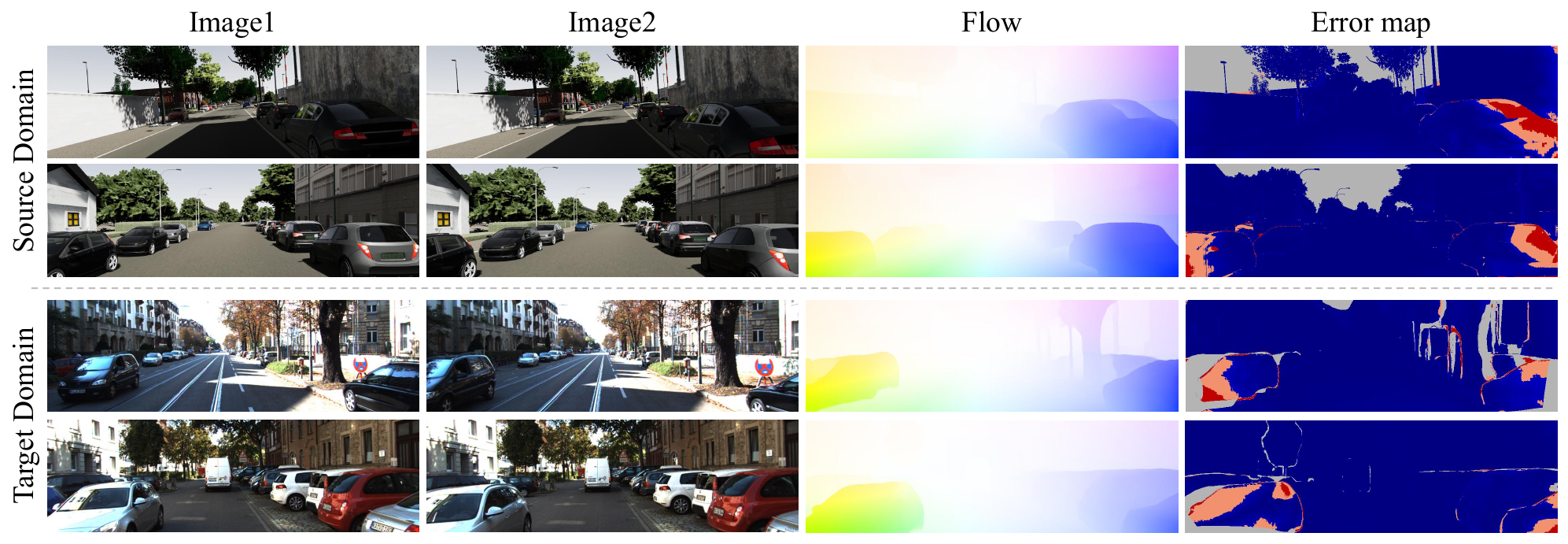}
    \vspace{-10pt}
    \caption{Visualization of error maps on the source and target domains. In the error maps, gray denotes non-evaluation, blue indicates errors less than $\mu + \sigma$, orange represents errors within the range of $\left [ \mu + \sigma ,  \mu + 3\sigma \right )$, and red signifies errors greater than or equal to $\mu + 3\sigma$.}
    \label{fig:acw}
    \vspace{-10pt}
\end{figure}

\subsection{Ablation Study}
\hspace{1em} Within the comprehensive FlowDA framework, we systematically deactivate individual modules to assess the significance of each component. We conduct our training on the Ktest dataset and perform testing on KITTI-2015's training set, with the results being summarized in Table~\ref{tab:ablation_main}. The findings reveal that the components ‘source domain supervision’, ‘crop’, and ‘unsupervised loss’ emerge as the most crucial, with their removal resulting in a reduction of overall performance by 15.4\%, 12.1\%, and 11.4\%, respectively.

‘w/o source domain supervision’ means that the source domain branch is removed. In this configuration, the model can be viewed as pre-trained in the source domain and then fine-tuned in a self-training and unsupervised manner in the target domain. When the source domain branch is removed during training, the model becomes prone to overfitting on noisy pseudo-labels and unsupervised loss, leading to poor results. In fact, at the end of the training, removing the source domain branch produces worse performance than what is reported in Table~\ref{tab:ablation_main}.

‘w/o crop’ means that the input images for the teacher model and the student model are in the same area, which results in a 12.1\% performance drop.

‘w/o unsupervised loss’ means to remove the loss function $L_{U}$. Photometric consistency loss is a characteristic of matching tasks such as optical flow estimation, and it represents the most significant difference between FlowDA and unsupervised domain adaptation methods used in segmentation, classification, and other fields. Photometric consistency loss can mine the supervision information of the target domain itself, and removal results in an 11.4\% performance degradation. 

‘w/o occ mask’ denotes that the occlusion mask $O_{t,c}$ is not employed to exclude the occluded region while computing the loss function $L_{A}$. Not using the occlusion mask to exclude areas where the pseudo-label is inaccurate results in a 7.4\% performance degradation.

‘w/o ACW’ refers to the exclusion of the Adaptive Curriculum Weighting module. The ACW module contributes to optimizing the overall training process and facilitates convergence. The visualization results in Figure~\ref{fig:acw} indicate that regions with significant learning challenges primarily correspond to out-of-frame occluded areas of foreground objects. Furthermore, the ACW module is beneficial for supervised approaches. Table~\ref{tab:ablation_acw} compares the results of supervised training in C and T with and without ACW. The findings indicate that ACW positively impacts the model’s ability to model in-domain data and generalize to cross-domain data.

‘w/o EMA’ means not using EMA's updated mean-teacher architecture, but using a single model for self-training. The experimental results show that the mean-teacher architecture updated by EMA can improve the stability of pseudo-labels.

‘w/o all’ refers to testing the optical flow model directly on KITTI-15 after training on C+T+V, serving as our baseline. Compared to the baseline, FlowDA exhibits a 36.0\% improvement in EPE and a 34.4\% improvement in Fl.

\begin{table}[t]
\centering
\setlength{\tabcolsep}{3.0mm}
\renewcommand{\arraystretch}{1.1}
\begin{tabular}{lcc}
\toprule[1.5pt]
\multirow{2}{*}{Experiment} & \multicolumn{2}{c}{KITTI-15(train)} \\ \cline{2-3} 
 & EPE & Fl-all \\ \hline
FlowDA & 1.60 & 5.27 \\
w/o source domain supervision & 2.10 & 6.08 \\
w/o crop & 1.98 & 5.91 \\
w/o unsupervised loss & 1.73 & 5.87 \\
w/o occ mask & 1.71 & 5.66 \\
w/o ACW & 1.64 & 5.51 \\
w/o EMA & 1.68 & 5.49 \\
w/o all & 2.50 & 8.03 \\ \hline
\end{tabular}
\setlength{\abovecaptionskip}{5pt}    
\setlength{\belowcaptionskip}{-10pt}
\vspace{-2pt}
\caption{Ablation study. Within the comprehensive FlowDA framework, individual modules are systematically deactivated to assess the significance of each component.}
\vspace{3pt}
\label{tab:ablation_main}
\end{table}

\begin{table}[t]
\centering
\setlength{\tabcolsep}{1.1mm}
\renewcommand{\arraystretch}{1.1}
\begin{tabular}{lcclcclcc}
\toprule[1.5pt]
\multirow{3}{*}{Method} & \multirow{3}{*}{ACW} & C &  & \multicolumn{5}{c}{C+T} \\ \cline{3-3} \cline{5-9} 
 &  & Chairs &  & \multicolumn{2}{c}{\begin{tabular}[c]{@{}c@{}}Sintel\\ (train)\end{tabular}} &  & \multicolumn{2}{c}{\begin{tabular}[c]{@{}c@{}}KITTI-15\\ (train)\end{tabular}} \\ \cline{3-3} \cline{5-6} \cline{8-9} 
 &  & EPE &  & Clean & Final &  & EPE & Fl-all \\ \hline
\multirow{2}{*}{RAFT} & \multicolumn{1}{l}{} & 0.829 &  & 1.43 & 2.71 &  & 5.04 & 17.4 \\
 & $\surd$ & 0.787 & \textbf{} & 1.27 & 2.70 & \textbf{} & 4.58 & 16.1 \\ \hline
\end{tabular}
\setlength{\abovecaptionskip}{5pt}    
\setlength{\belowcaptionskip}{-10pt}
\vspace{-2pt}
\caption{Effectiveness of ACW module in supervised learning.}
\vspace{0pt}
\label{tab:ablation_acw}
\end{table}

\subsection{FlowDA Performance}
\subsubsection{Comparison with Unsupervised Domain Adaptation Methods}
\hspace{1em} Due to the scarcity of unsupervised domain adaptation methods for optical flow, we only select StereoFlowGAN~\cite{xiong2023stereoflowgan} and FlowSupervisor~\cite{im2022semi} for comparison. When comparing with FlowSupervisor, we use Ktest for training and report the results on the KITTI-2015~\cite{menze2015joint} training set. When comparing with StereoFlowGAN, we follow its protocol, using only the KITTI-2015 training set's 200 image pairs and the VKITTI~\cite{cabon2020virtual} dataset, not pre-trained on other datasets. For the test set, we select one pair for every five pairs from the 200 pairs of images within the KITTI-2015 training set, resulting in a total of 40 pairs of images. The rest is used as a training set, which does not use optical flow labels. In addition, we also compare with MIC~\cite{hoyer2023mic}, the most advanced domain adaptive method in classification, segmentation, and other fields. We adapt MIC to optical flow estimation and maintain the same training and evaluation strategy.

\begin{table}[t]
\centering
\setlength{\tabcolsep}{1.7mm}
\renewcommand{\arraystretch}{1.1}
\begin{tabular}{lcclcc}
\toprule[1.5pt]
\multirow{2}{*}{Method} & \multicolumn{2}{c}{\begin{tabular}[c]{@{}c@{}}KITTI-15\\ (train-split)\end{tabular}} &  & \multicolumn{2}{c}{\begin{tabular}[c]{@{}c@{}}KITTI-15\\ (train)\end{tabular}} \\ \cline{2-3} \cline{5-6} 
 & EPE & Fl-all &  & EPE & Fl-all \\ \hline
StereoFlowGAN~\cite{xiong2023stereoflowgan} & 5.19 & 18.32 &  & - & - \\
FlowSupervisor~\cite{im2022semi} & - & - &  & 2.39 & 7.63 \\
MIC~\cite{hoyer2023mic} & 2.43 & 6.67 &  & 2.10 & 6.56 \\ \hline
FlowDA(Ours) & \textbf{1.85} & \textbf{5.36} &  & \textbf{1.60} & \textbf{5.27} \\ \hline
\end{tabular}
\setlength{\abovecaptionskip}{5pt}    
\setlength{\belowcaptionskip}{-10pt}
\vspace{-2pt}
\caption{Comparison with unsupervised domain adaptation methods. The best results are shown in bold.}
\vspace{3pt}
\label{tab:comparison_uda}
\end{table}

\begin{table}[t]
\centering
\setlength{\tabcolsep}{2.0mm}
\renewcommand{\arraystretch}{1.1}
\begin{tabular}{clcclc}
\toprule[1.5pt]
\multirow{2}{*}{} & \multirow{2}{*}{Method} & \multicolumn{2}{c}{\begin{tabular}[c]{@{}c@{}}KITTI-15\\ (train)\end{tabular}} &  & \begin{tabular}[c]{@{}c@{}}KITTI-15 \\ (test)\end{tabular} \\ \cline{3-4} \cline{6-6} 
 &  & EPE & Fl-all &  & Fl-all \\ \hline
\multirow{3}{*}{MF} & SelFlow~\cite{liu2019selflow} & 4.84 & - &  & 14.19 \\
 & ARFlow~\cite{liu2020learning} & 3.46 & - &  & 11.79 \\
 & SMURF~\cite{stone2021smurf} & 2.01 & 6.72 &  & 6.83 \\ \hline
\multirow{3}{*}{TF} & UFlow~\cite{jonschkowski2020matters} & 2.84 & 9.39 &  & 11.13 \\
 & SMURF*~\cite{stone2021smurf} & 2.45 & 7.53 &  & - \\ \cline{2-6} 
 & FlowDA(Ours) & \textbf{1.60} & \textbf{5.27} & \textbf{} & \textbf{5.16} \\ \hline
\end{tabular}
\setlength{\abovecaptionskip}{5pt}    
\setlength{\belowcaptionskip}{-10pt}
\vspace{-2pt}
\caption{Comparison with unsupervised methods. Methods that only use two frames are denoted with ‘TF’, while methods that use multiple frames in training or testing are denoted with ‘MF’. ‘-’ indicates no results reported. ‘*’ denotes the two-frame version of the corresponding method. The best results are shown in bold.}
\vspace{0pt}
\label{tab:comparison_un}
\end{table}

Table~\ref{tab:comparison_uda} reveals a significant improvement in our method, outperforming StereoFlowGAN~\cite{xiong2023stereoflowgan} and FlowSupervisor~\cite{im2022semi} by 70.7\% and 30.9\% on Fl under identical settings, respectively. While MIC~\cite{hoyer2023mic} is recognized as an exceptional unsupervised domain adaptive framework, its performance is restricted due to its lack of consideration for the specific characteristics of optical flow estimation. Our FlowDA outperforms MIC by a substantial margin, achieving a 19.7\% improvement on Fl.

\subsubsection{Comparison with Unsupervised Methods}
\hspace{1em} Table~\ref{tab:comparison_un} shows the comparison of FlowDA to the most advanced unsupervised methods, both two-frame and multi-frame. For the KITTI-2015 training set results, we use Ktest for training, and for the KITTI-2015 test set results, we use Ktrain for training.

FlowDA outperforms state-of-the-art unsupervised methods by efficiently utilizing labeled virtual datasets and unlabeled real datasets. Compared to the most advanced two-frame unsupervised approach~\cite{stone2021smurf}, FlowDA achieves a 30.0\% improvement on Fl and 34.7\% improvement on EPE of the KITTI-2015 training set. In comparison to the state-of-the-art multi-frame unsupervised method SMURF~\cite{stone2021smurf}, FlowDA performs 21.6\% better of Fl on the KITTI-2015 training set and 24.5\% better of Fl on the test set. These findings underscore the necessity of synthesizing virtual datasets to significantly enhance the performance of real-world optical flow models. In addition, it is important to note that SMURF~\cite{stone2021smurf} necessitates complex training with multiple stages for ensuring training stability, while FlowDA only requires a simple setup.

\begin{table}[t]
\centering
\setlength{\tabcolsep}{2.3mm}
\renewcommand{\arraystretch}{1.1}
\begin{tabular}{lcclcc}
\toprule[1.5pt]
\multirow{2}{*}{Method} & \multicolumn{2}{c}{\begin{tabular}[c]{@{}c@{}}KITTI-12\\ (train)\end{tabular}} &  & \multicolumn{2}{c}{\begin{tabular}[c]{@{}c@{}}KITTI-15\\ (train)\end{tabular}} \\ \cline{2-3} \cline{5-6} 
 & EPE & Fl-all &  & EPE & Fl-all \\ \hline
Depthstillation~\cite{aleotti2021learning} & 1.77 & 5.97 &  & 3.99 & 13.34 \\
RealFlow~\cite{han2022realflow} & 1.32 & 5.41 &  & 2.31 & 8.65 \\
MPI-Flow~\cite{liang2023mpi} & 1.24 & 4.51 &  & 2.16 & 7.30 \\ \hline
FlowDA(Ours) & \textbf{1.14} & \textbf{3.43} & \textbf{} & \textbf{1.60} & \textbf{5.27} \\ \hline
\end{tabular}
\setlength{\abovecaptionskip}{5pt}    
\setlength{\belowcaptionskip}{-10pt}
\vspace{-2pt}
\caption{Comparison with dataset generation methods. The best results are shown in bold.}
\vspace{0pt}
\label{tab:comparison_dg}
\end{table}

\subsubsection{Comparison with Dataset Generation Methods}
\hspace{1em} We compare all the methods for generating real-world datasets, including Depthstillation~\cite{aleotti2021learning}, RealFlow~\cite{han2022realflow}, and MPI-Flow~\cite{liang2023mpi}, and the results are summarized in Table~\ref{tab:comparison_dg}. These methods often require additional models to assist in the generation of datasets. Specifically, Depthstillation~\cite{aleotti2021learning} and RealFlow~\cite{han2022realflow} require pre-trained depth estimation models, while MPI-Flow~\cite{liang2023mpi} also requires segmentation models in addition to depth estimation models. We follow the procedures of these methods, using the KITTI-2015 multi-view test set as the target domain dataset, and report the results on the KITTI-2012 training set and the KITTI-2015 training set.

Compared with MPI-Flow~\cite{liang2023mpi}, the best method for generating real datasets, our method FlowDA improves Fl by 23.9\% and 27.8\% in KITTI-2012 and KITTI-2015, respectively, demonstrating the potential to improve the performance of optical flow models in the real world by synthesizing virtual datasets with accurate labels. Figure~\ref{fig:vis_kitti} shows qualitative results on KITTI. FlowDA achieves significant improvement in shadow areas and large reflective areas.

\begin{figure}[t]
    \centering
    \includegraphics[width=1.0\linewidth]{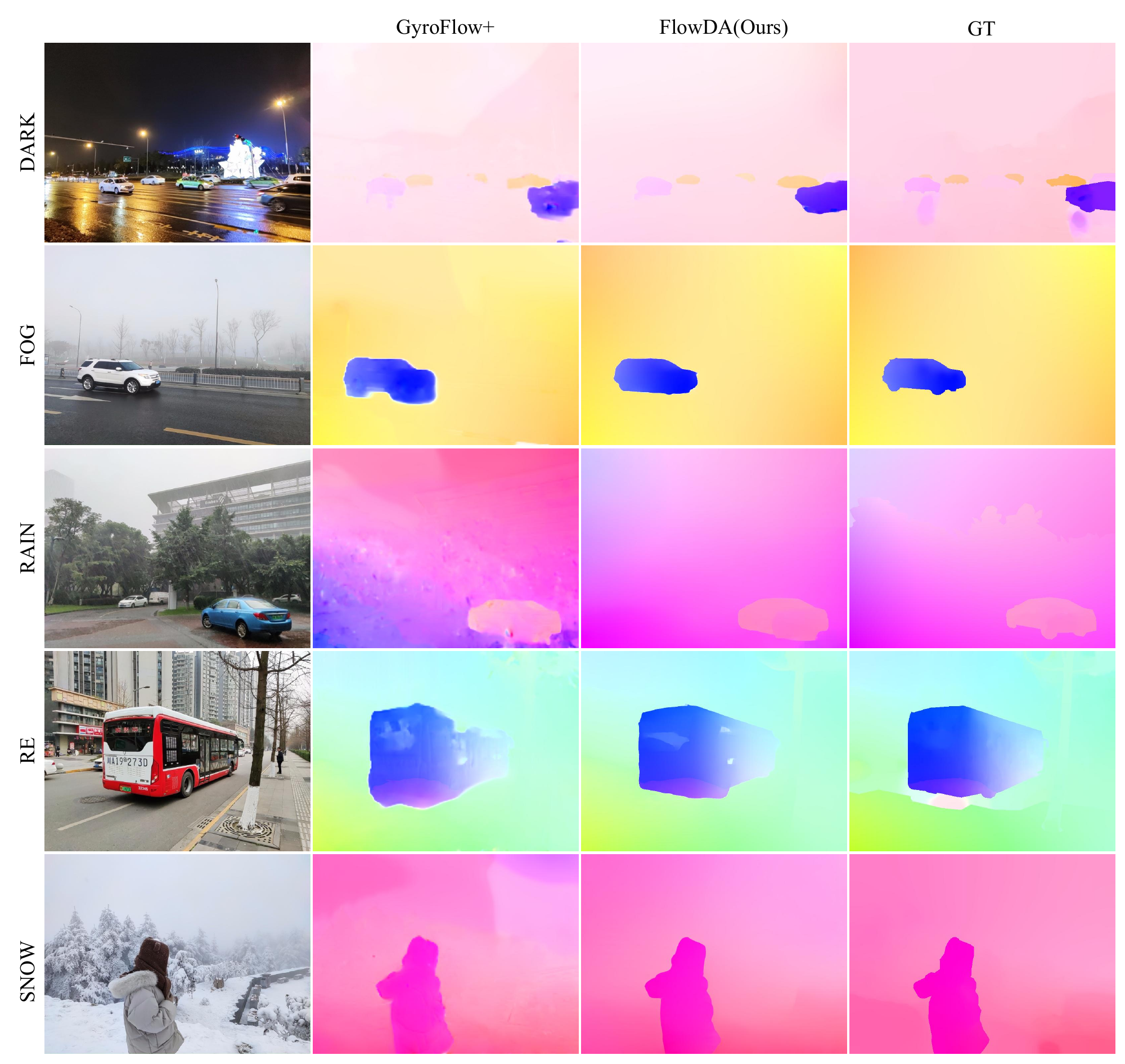}
    \vspace{-12pt}
    \caption{Visual comparison on GHOF evaluation set. The first to fifth lines are different scenarios. The second to third columns represent the results of GyroFlow+~\cite{li2023gyroflow+} and our FlowDA. The fourth column is the Ground Truth.}
    \label{fig:vis_ghof}
    \vspace{-15pt}
\end{figure}

\subsection{Versatility of FlowDA}

\hspace{1em} This section aims to verify the effectiveness of FlowDA in various weather conditions and scenarios. To determine the suitable source domain, several pre-trained models are tested on the GHOF test set, and the results are summarized in Table~\ref{tab:ghof_test}. Surprisingly, model pre-trained on Flyingthings~\cite{mayer2016large} outperforms those pre-trained on VKITTI~\cite{cabon2020virtual} and KITTI~\cite{menze2015joint}. It is important to note that the GHOF~\cite{li2023gyroflow+} dataset consists of real scenarios. This is partly due to the disparities in scene content between GHOF, VKITTI, and KITTI datasets. Another notable difference lies in their motion patterns. Compared to random object movement in Flyingthings, VKITTI is much closer to the real world. However, there are notable differences between VKITTI and GHOF datasets in terms of motion patterns, as VKITTI is an autonomous dataset with a camera positioned in front of the vehicle while GHOF is captured using a handheld phone. At this point, pre-training on the Flyingthings, which contains various random movements, performs better on the GHOF, and even better than pre-training on the real autonomous driving dataset KITTI. We surmise that the motion pattern holds more significance than image content and style in measuring the domain gap in optical flow estimation.

Based on the above findings, we use Flyingthings~\cite{mayer2016large} as the source domain data, the training set of GHOF as the target domain data, and evaluate on the test set of GHOF. The results can be seen in Table~\ref{tab:comparison_ghof}. FlowDA demonstrates a 21.0\% improvement in average EPE compared to the next best method, GyroFlow+~\cite{li2023gyroflow+}. Visual comparisons are shown in Figure~\ref{fig:vis_ghof}. FlowDA exhibits more stable optical flow estimation results in a variety of scenarios, with clearer and sharper edges. It is noteworthy that in low-light scenarios, our FlowDA exhibits a slightly inferior performance compared to GyroFlow+, which combines gyroscope data to estimate the homography matrix and optical flow. This may be due to two reasons: (1) The difference between the source domain and the target domain is too large, resulting in FlowDA not showing complete performance. (2) In low-light scenes, the photometric consistency is greatly affected, and the use of scene-independent gyroscope data by GyroFlow+~\cite{li2023gyroflow+} can improve performance.

\begin{table}[t]
\centering
\setlength{\tabcolsep}{0.7mm}
\renewcommand{\arraystretch}{1.0}
\begin{tabular}{lcccc}
\toprule[1.5pt]
Dataset & Chairs~\cite{dosovitskiy2015flownet} & Things~\cite{mayer2016large} & VKITTI~\cite{cabon2020virtual} & KITTI~\cite{menze2015joint} \\ \hline
EPE & 1.79 & 1.26 & 5.66 & 4.67 \\
Fl & 8.65 & 6.88 & 16.12 & 14.20 \\ \hline
\end{tabular}
\setlength{\abovecaptionskip}{5pt}    
\setlength{\belowcaptionskip}{-10pt}
\vspace{-2pt}
\caption{Comparison of results of pre-training weights on GHOF test sets. Average Fl and EPE across multiple scenes are reported.}
\vspace{3pt}
\label{tab:ghof_test}
\end{table}

\begin{table}[t]
\centering
\setlength{\tabcolsep}{1.0mm}
\renewcommand{\arraystretch}{1.0}
\begin{tabular}{lcccccc}
\toprule[1.5pt]
\multirow{2}{*}{Method} & \multicolumn{5}{c}{Scenes} & \multirow{2}{*}{Avg} \\ \cline{2-6}
 & Dark & Fog & Rain & RE & SNOW &  \\ \hline
Farneback~\cite{farneback2003two} & 7.11 & 5.18 & 1.80 & 2.40 & 4.79 & 4.15 \\
ARFlow~\cite{liu2020learning} & 5.36 & 2.83 & 0.99 & 2.07 & 4.90 & 3.23 \\
UFlow~\cite{jonschkowski2020matters} & 3.31 & 1.28 & 0.66 & 1.08 & 1.95 & 1.66 \\
UPFlow~\cite{luo2021upflow} & 3.45 & 1.20 & 0.54 & 1.06 & 1.88 & 1.62 \\
GyroFlow+~\cite{li2023gyroflow+} & \textbf{2.20} & 1.14 & 0.48 & 1.04 & 1.12 & 1.19 \\ \hline
FlowDA(Ours) & 2.38 & \textbf{0.52} & \textbf{0.40} & \textbf{0.55} & \textbf{0.85} & \textbf{0.94} \\ \hline
\end{tabular}
\setlength{\abovecaptionskip}{5pt}    
\setlength{\belowcaptionskip}{-10pt}
\vspace{-2pt}
\caption{Comparison on GHOF. EPE under different scenes are reported. The best results are shown in bold.}
\vspace{0pt}
\label{tab:comparison_ghof}
\end{table}

\section{Discussion}
\label{sec:discussion}

\hspace{1em} This paper proposes FlowDA, which demonstrates the effectiveness of utilizing virtual datasets to enhance the performance of optical flow estimation models in real-world scenarios. However, there are still certain aspects that require further exploration. Firstly, a clear definition of the measurement standard for domain gap in optical flow datasets is lacking, which can guide the selection and synthesis of source domain virtual datasets for practical applications. Secondly, the performance of low-light scenes remains subpar even after domain adaptation. In addition to the lack of relevant datasets, low-light scenes are more challenging due to their characteristics and may require some specialized design or even multimodal data.

\section{Conclusion}
\label{sec:conclusion}
\hspace{1em} The expensive nature of labeling optical flow makes it challenging to gather and construct real-world optical flow datasets. The limited availability of datasets significantly hinders the performance of optical flow models in real-world scenarios. Virtual datasets offer a viable alternative, with effective domain adaptation being the key element. In this paper, we introduce FlowDA, a domain adaptive framework for optical flow estimation. By reasonably combining the mean-teacher-based UDA architecture, unsupervised optical flow estimation techniques, and Adaptive Curriculum Weighting module, FlowDA significantly enhances the performance of optical flow estimation models in real-world scenarios. We hope that our work will provide robust support and inspiration for the practical application of optical flow models in real-world settings.

\clearpage
{
    \small
    \bibliographystyle{ieeenat_fullname}
    \bibliography{main}
}


\end{document}